\documentclass{article}
\usepackage{spconf,amsmath,graphicx}
\usepackage{adjustbox}
\title{Towards Light Weight Object Detection System}

\name{
\begin{tabular}{c}
 Dharma KC$^{1,\star}$\thanks{$^{\star}$Work performed during internship at Qualcomm Technologies, Inc.}, Venkata Ravi Kiran Dayana$^{2}$, Meng-Lin Wu$^{2}$\\
 Venkateswara Rao Cherukuri$^{\dagger}$\thanks{$^{\dagger}$Work performed while at Qualcomm Technologies, Inc.}, Hau Hwang$^{2}$
\end{tabular}}
\address{$^{1}$Department of Computer Science, University of Arizona\\
$^{2}$Qualcomm Technologies, Inc.}

\begin{document}
\maketitle
\begin{abstract}
    Transformers are a popular choice for classification tasks and as backbones for object detection tasks. However, their high latency brings challenges in their adaptation to light weight object detection systems. We present an approximation of the self-attention layers used in the transformer architecture. This approximation reduces the latency of the classification system while incurring minimal loss in accuracy. We also present a method that uses a transformer encoder layer for multi-resolution feature fusion. This feature fusion improves the accuracy of the state-of-the-art light weight object detection system without significantly increasing the number of parameters. Finally, we provide an abstraction for the transformer architecture called Generalized Transformer (gFormer) that can guide the design of novel transformer-like architectures.
\end{abstract}
\begin{keywords}
Vision transformer, self-attention, object detection, deep neural networks
\end{keywords}

\section{Introduction}
Convolutional neural networks (CNNs)~\cite{lecun1995convolutional} have been widely used as backbones for object detection systems. MobileNets~\cite{howard2017mobilenets} use depthwise separable convolutions to develop light weight CNNs. MobileNetV2 further improves MobileNets using inverted residuals and linear bottlenecks. It also introduced efficient ways of applying depthwise separable convolutions to the heads of Single Shot MultiBox Detector (SSD)~\cite{liu2016ssd}, which resulted in the light weight object detection system, SSDLite. Recently, Vision Transformers (ViTs)~\cite{dosovitskiy2021an} are gaining popularity due to their ability to extract global information. However, they lack the spatial inductive biases present in CNNs. MobileViT~\cite{mehta2022mobilevit} presented a hybrid architecture based on CNNs and ViTs that leverages the inductive biases of CNNs and also includes global information through ViTs. MobileViT achieves impressive performance on the ImageNet-1k classification dataset~\cite{russakovsky2015imagenet}, and its main disadvantage is high latency.

In this work, we propose \emph{Convolution as Transformer} (CAT): a module that approximates the self-attention layer in transformers. CAT has low latency and thus can be used in light weight systems for image classification and object detection. We replace expensive transformer blocks used in MobileViT with our CAT blocks, and we show that they are competitive with the self-attention modules for image classification tasks. Moreover, CAT blocks have complexity $O(n\times d)$, unlike self-attention that has complexity $O(n^2 \times d)$, where $n$ is the sequence length, and $d$ is feature vector size.

Existing light weight systems for object detection~\cite{mehta2022mobilevit, liu2016ssd} mainly consist of a backbone to extract features from images, followed by heads to extract features from multiple output resolutions. Predictions on object label and localization are made directly from these multi-scale features. It is therefore challenging to learn the relationship between these features from multiple scales, carrying different semantic information.

To overcome this, we propose the module \emph{Transformer Encoder as Feature Fusion} (TAFF): a single layered transformer encoder~\cite{vaswani2017attention} which fuses features from multiple resolutions at different scales. We show empirically that the feature fusion performed by TAFF improves the accuracy of state-of-the-art object detection models like MobileViT~\cite{mehta2021mobilevit}. 

Finally, we propose \emph{Generalized TransFormer} (gFormer): a general abstract architecture that binds multiple variations of attention and transformer mechanisms under a common umbrella. From this perspective, MetaFormer~\cite{yu2022metaformer}, Transformer~\cite{vaswani2017attention}, Squeeze and Excitation Networks~\cite{hu2018squeeze,tolstikhin2021mlp}, and our CAT block are all variations of gFormer.

\section{System}
\subsection{Convolution as Transformer (CAT)}
\begin{figure*}[h]
    \centering
    \includegraphics[width=\textwidth, height=2.5in]{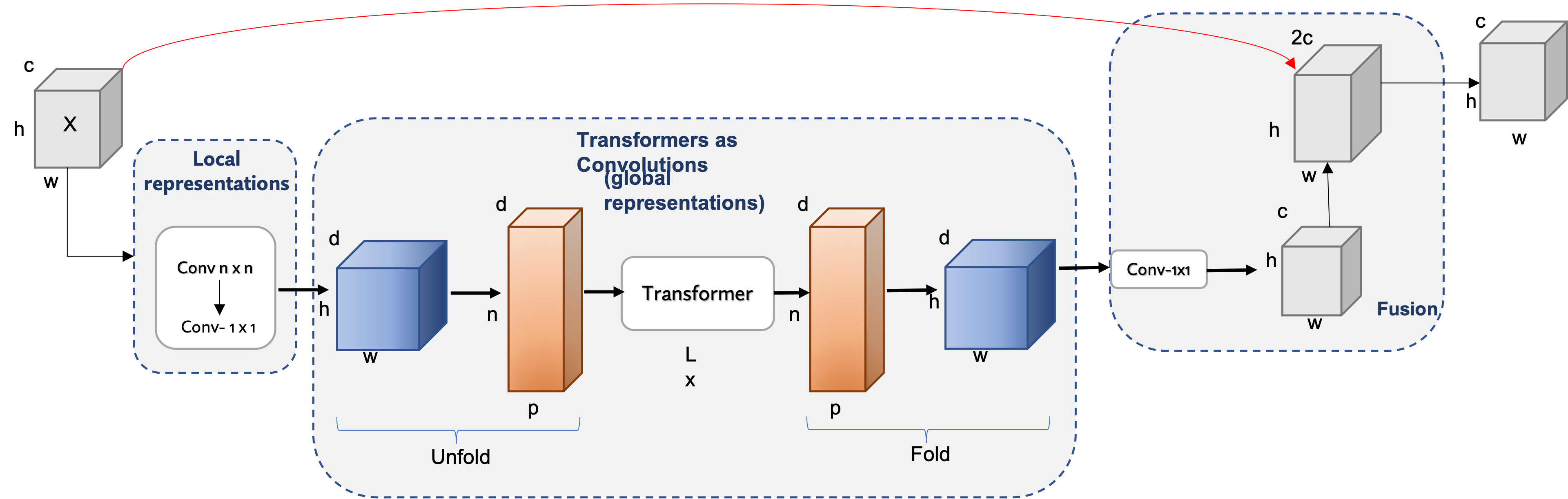}
    \caption{MobileViT block used in the MobileVit~\cite{mehta2021mobilevit}}
    \label{fig:mobilevit_block}
\end{figure*}

The baseline for this architecture is the MobileViT architecture ~\cite{mehta2021mobilevit} that uses MobileNetV2 blocks along with MobileViT blocks that contain transformer layers for extracting global information. We refer to ~\cite{mehta2021mobilevit} for the full architecture and only show the MobileViT block in Fig.~\ref{fig:mobilevit_block}.

\begin{figure*}[h]
    \centering
    \includegraphics[width=\textwidth, height=2.5in]{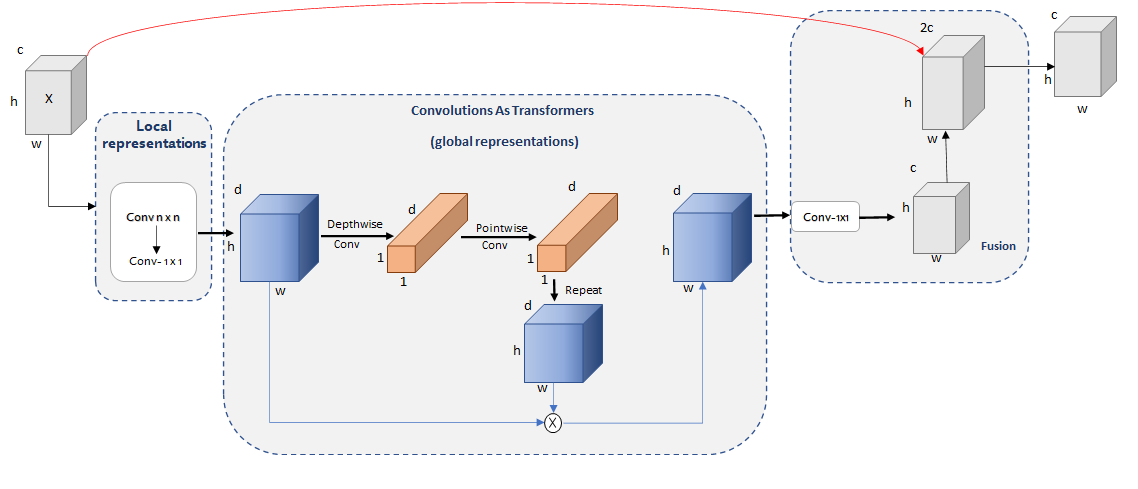}
    \caption{Convolution As a Transformer (CAT) block}
    \label{fig:cat_block}
\end{figure*}
The MobileViT architecture extracts global information with transformers. The major disadvantage of the above method is that it has high latency because of the self-attention layers used inside the transformer as convolutions block. We hypothesize and prove empirically that we can extract the global information using simpler functions (with slight loss in accuracy) that has lower latency and lower number of computations. Thus, we propose Convolution as Transformer (CAT) blocks (Fig.~\ref{fig:cat_block}) that approximates this self-attention for global feature extraction. We propose to use the following function to approximate transformers by convolution blocks:

\begin{itemize}
    \item Depthwise separable convolutional filter to extract global information:\\
        \texttt{global\_information} == \texttt{dep\_sep\_conv(x)}
        This is a combination of depthwise convolution with kernel size of (h, w) and a pointwise convolution. This can also be interpreted as a combination of spatial MLP and channel MLP where spatial MLP extracts information from spatial domain and channel MLP extracts information from channel dimension. Thus, the output vector represents approximate global information from spatial and channel dimension.
    \item Broadcast global information to same shape as intermediate feature map:\\
        y = \texttt{reshape(global\_information, (H, W, D))}
    \item Elemenentwise product between global information and intermediate feature map:\\
        x = \texttt{elementwise\_product(x, y)}. Here, $\otimes$ represents elementwise product between two tensors. 
\end{itemize}

\subsection{Transformer Encoder as Feature Fusion (TAFF)}
Fig.~\ref{fig:taff} shows the architecture of our TAFF block:
\begin{figure*}[h]
    \centering
    \includegraphics[width=0.9\textwidth, height=1.8in]{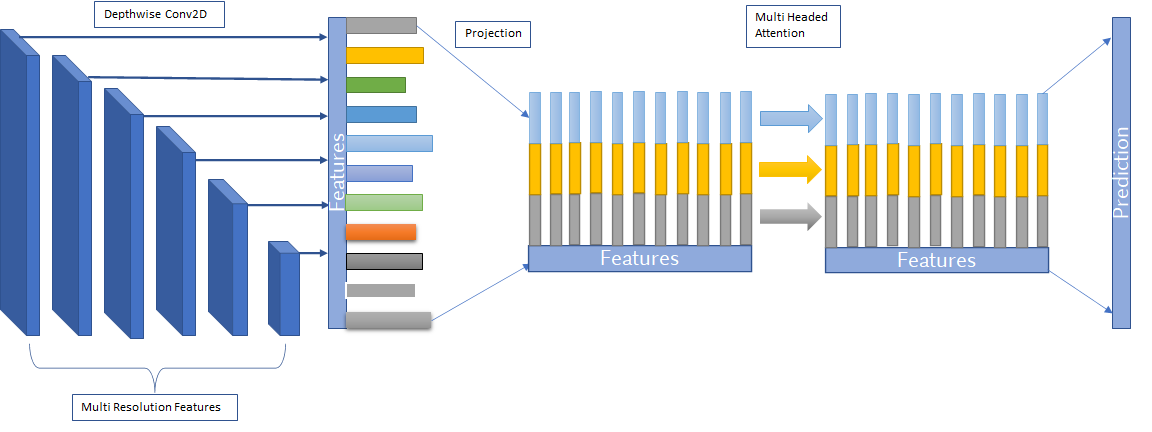}
    \caption{TAFF module for feature fusion from multiple scales}
    \label{fig:taff}
\end{figure*}
We extract features from multiple resolution intermediate feature maps. We extract $d$ dimensional features from each anchor box from multiple resolution feature maps, we then use Transformer encoder layer that first projects these features $n * d$ feature tensor into key, query and values tensors. It then applies a multi-headed attention to these features giving each network head an ability to not only look at the feature vector at the corresponding location but also look at other feature vectors to make an informed decision. For example the locations at the lower level can look at the features of the semantically higher layers to make a good prediction.

\subsection{gFormer}
MetaFormer~\cite{yu2022metaformer} abstracts different variations of the transformer architectures into a general framework. We further generalize multiple variations of the attention architecture into a common framework that even includes the squeeze and excite networks. Fig.~\ref{fig:gformer} shows the general transformer (gFormer) architecture:
\begin{figure}[h]
\centering
\begin{minipage}{.5\textwidth}
  \centering
  \includegraphics[width=.4\linewidth, height=2in]{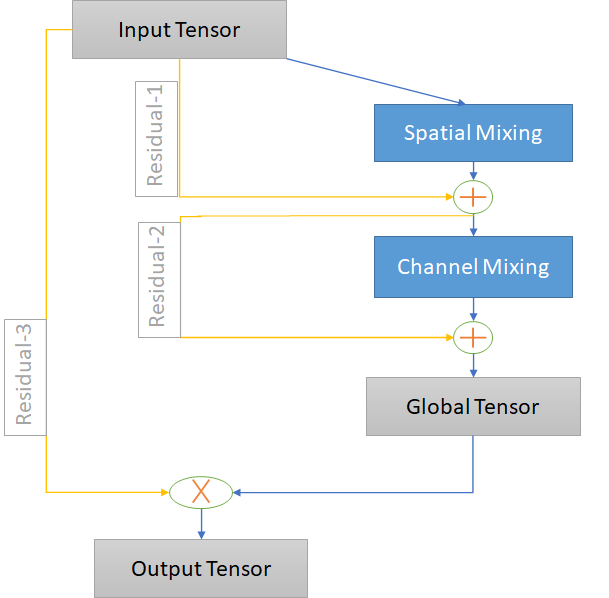}
  \caption{gFormer: generalized abstraction for transformer networks}
  \label{fig:gformer}
\end{minipage}
\end{figure}

Where $\oplus$ represents summation, and $\otimes$ represents any pairwise interaction function (e.g. elementwise product). Thus, following architectures become special cases of the gFormer abstraction:
\begin{itemize}
    \item Transformer~\cite{vaswani2017attention}: Transformer captures pairwise interaction using multi-headed attention, so we don't need residual-3 and last $\otimes$. Thus, if we remove residual-3, use multi-headed attention for spatial mixing, and use MLP for channel mixing with normalization layers, we recover the Transformer architecture.
    
    \item MetaFormer~\cite{yu2022metaformer}: If we remove residual-3 and pairwise interaction function we obtain the MetaFormer.

    \item CAT: If we remove residual-1 and residual-2, use depthwise convolution or spatial MLP for spatial mixing, pointwise convolution or MLP for channel mixing, repeat the global vector to input tensor shape and take the Hadamard product between two tensors, we arrive at our CAT block.
    \item Squeeze and excite networks~\cite{hu2018squeeze,tolstikhin2021mlp}: If we remove residual-1, residual-2, use global pooling as spatial mixing, use a linear layer with "MLP-ReLU-MLP-sigmoid" as channel mixing, repeat the output of this linear layer to original input tensor shape and take Hadamard product, we obtain squeeze and excite networks.
    \item MLP-Mixer~\cite{tolstikhin2021mlp}: If we remove residual-3, and pairwise interaction function, use MLP for spatial mixing and MLP for channel mixing, we obtain the MLP-Mixer.
\end{itemize}

The gFormer abstraction helps us create new architectures that are optimized for different needs and still reap the benefits of transformers for capturing global information.

\section{Experiments}

\subsection{Classification with CAT}
For classification experiments, we use a subset of ImageNet-1k~\cite{russakovsky2015imagenet} dataset which consists of 100 balanced classes from ImageNet-1k keeping the original train-validation split. We used a subset of the dataset because of computational overhead of training on whole ImageNet-1k. The training dataset consists of 130k images, while the validation dataset consists of 5k images. We follow~\cite{mehta2022mobilevit} and report accuracy on the validation dataset. We used PyTorch~\cite{paszke2019pytorch} for our experiments. We trained our model up to 300 epochs with the stochastic gradient descent (SGD) algorithm with weight decay of $4e-5$ and momentum of 0.5, at a batch size of 64 per GPU on 4 Nvidia RTX 2080Ti GPUs. We used the Cosine Annealing scheduler with an initial learning rate of 0.05, increasing to 0.4 within 7500 iterations, and ultimately decreasing to a minimum of  $2e-4$. We use an input image resolution of $256\times256$. We used the Swish activation function~\cite{ramachandran2017searching} as our default activation function.

We compare our results with MobileViT as a baseline method. We also modified the token mixing algorithm based on Fourier transforms called FNet~\cite{51420} architecture used in natural language processing domain and replaced the MobileViT block with this mechanism. Note that we replaced the multiple transformer layers used in MobileViT with a single layer of our CAT block. Also, our CAT block doesn't use any feedforward layers used by the transformer layers in MobileViT.

\subsection{Detection with TAFF}
We use COCO~\cite{lin2014microsoft} dataset for object detection experiments with the same setting and same hyperparameters as MobileViT~\cite{mehta2022mobilevit}. The input image resolution is $320\times320$. We evaluated the performance on the validation dataset using mAP@IOU 0.50:0.05:0.95. We first train MobileViT with SSDLite as the baseline method.

We then extract features from MobileViT and SSDLite and fuse features from multiple layers using our TAFF module. The fused features are used to make the prediction for the class and bounding box. Note that we didn't even use positional encoding for the feature fusion that helps in keeping the small memory footprint when deployed on mobile devices. It's interesting that the single layer of transformer encoder was able to fuse features from multiple scales and multiple anchor boxes. We hypothesize that the projection matrices in the transformer encoder project features from multiple scales and anchor boxes to a common domain suitable for feature fusion.

\section{Results}
\subsection{Classification with CAT}
Table~\ref{table:classification} shows the accuracy, number of parameters (in Million), number of floating point operations (FLOPS), and latency (measured for a single image on Nvidia RTX 2080 Ti by averaging over 1000 runs). The results show that our CAT block closely approximates the self-attention module of the MobileViT architecture with slight decrease in accuracy but at almost half number of parameters, half FLOPs, and half latency.

We also modified the FNet architecture~\cite{51420} used in the natural language processing (NLP) domain to approximate the self-attention layers in the MobileViT architecture and obtained good results. This demonstrates that FNet is also a viable approach for computer vision along with NLP domain.

\vspace{-3mm}

\begin{table}[h]
\caption{CAT classification performance on ImageNet dataset}
    \centering
    \begin{adjustbox}{width=\columnwidth,height=0.4in,center}
        \begin{tabular}{ |c|c|c|c|c|c| } 
         \hline
         Model & Accuracy & Parameters & FLOPS & Latency & Complexity \\ 
         \hline
         MobileViT  & 84.84 & 5.7 M & 4 G & 11.17 ms & $O(n^2d)$ \\ 
         \hline
         $\textbf{CAT}$ &  $\textbf{83.84}$ & $\textbf{2.4 M}$ & $\textbf{1.27 G}$ & $\textbf{4.69 ms}$ & $\textbf{$O(nd)$}$\\ 
         \hline
         FNet &  84.34 & 2 M & 1.25 G & 3.87 ms & $O(nlog(n)d)$\\
         \hline
        \end{tabular}
        \end{adjustbox}
        \label{table:classification}
\end{table}

Table~\ref{table:classification} also shows the advantage of our method compared to other methods in terms of run-time complexity of self-attention module. We can see that our method has linear complexity $O(n)$ with respect to sequence length, while other methods have either $O(nlogn)$ or $O(n^2)$ complexity.

\subsection{Detection with TAFF}
Table~\ref{table:detection} shows the effectiveness of our TAFF module in object detection tasks on the COCO validation dataset, which is a challenging object detection dataset. We demonstrate that adding this module on top of the SSD architecture with the MobileViT backbone significantly increases the accuracy of object detection by about 2.5mAP on the COCO validation dataset, while the number of parameters only grew by $7\%$

\vspace{-3mm}

\begin{table}[h]
\caption{TAFF detection performance on COCO dataset}
\begin{center}
\begin{tabular}{ |c|c|c| }
 \hline
 Method & mAP & Parameters\\ 
 \hline
 MobileNetv1 & 22.2 & 5.1 M \\
 \hline 
 MobileNetv2 & 22.1 & 4.3 M \\
 \hline
 MobileNetv3 & 22.0 & 4.9 M \\
 \hline
 MobileViT & 27.7 & 5.7 M  \\ 
 \hline
 $\textbf{TAFF}$ & $\textbf{30.1}$ & $\textbf{6.1 M}$\\
 \hline
\end{tabular}
\end{center}
\label{table:detection}
\end{table}

\vspace{-8mm}

\section{Conclusion}
In this paper, we have proposed the CAT block that can decrease the latency and FLOPs of transformer-based backbones such as MobileViT, while increasing the inference speed using simple approximation functions composed of depthwise-separable convolution and Hadamard product. Then, we have proposed the TAFF module that improves the accuracy of existing anchor-based object detection systems by fusing features from multiple scales. Finally, we have presented a general framework called gFormer that helps us design new architectures.

There are multiple ways to combine these ideas with existing systems to develop further light weight, accurate, and fast object detection systems. One such approach would be fusing these ideas to anchor-free object detection systems. For example, our CAT block and TAFF module can be integrated with DETR~\cite{carion2020end}. We explore this approach in future research.

\vfill\pagebreak

\clearpage
\bibliographystyle{IEEEbib}
\bibliography{main}

\begin{thebibliography}{10}

\bibitem{lecun1995convolutional}
Yann LeCun, Yoshua Bengio, et~al.,
\newblock ``Convolutional networks for images, speech, and time series,''
\newblock {\em The handbook of brain theory and neural networks}, vol. 3361,
  no. 10, pp. 1995, 1995.

\bibitem{howard2017mobilenets}
Andrew~G Howard, Menglong Zhu, Bo~Chen, Dmitry Kalenichenko, Weijun Wang,
  Tobias Weyand, Marco Andreetto, and Hartwig Adam,
\newblock ``Mobilenets: Efficient convolutional neural networks for mobile
  vision applications,''
\newblock {\em arXiv preprint arXiv:1704.04861}, 2017.

\bibitem{liu2016ssd}
Wei Liu, Dragomir Anguelov, Dumitru Erhan, Christian Szegedy, Scott Reed,
  Cheng-Yang Fu, and Alexander~C Berg,
\newblock ``Ssd: Single shot multibox detector,''
\newblock in {\em European conference on computer vision}. Springer, 2016, pp.
  21--37.

\bibitem{dosovitskiy2021an}
Alexey Dosovitskiy, Lucas Beyer, Alexander Kolesnikov, Dirk Weissenborn,
  Xiaohua Zhai, Thomas Unterthiner, Mostafa Dehghani, Matthias Minderer, Georg
  Heigold, Sylvain Gelly, Jakob Uszkoreit, and Neil Houlsby,
\newblock ``An image is worth 16x16 words: Transformers for image recognition
  at scale,''
\newblock in {\em International Conference on Learning Representations}, 2021.

\bibitem{mehta2022mobilevit}
Sachin Mehta and Mohammad Rastegari,
\newblock ``Mobilevit: Light-weight, general-purpose, and mobile-friendly
  vision transformer,''
\newblock in {\em International Conference on Learning Representations}, 2022.

\bibitem{russakovsky2015imagenet}
Olga Russakovsky, Jia Deng, Hao Su, Jonathan Krause, Sanjeev Satheesh, Sean Ma,
  Zhiheng Huang, Andrej Karpathy, Aditya Khosla, Michael Bernstein, et~al.,
\newblock ``Imagenet large scale visual recognition challenge,''
\newblock {\em International journal of computer vision}, vol. 115, no. 3, pp.
  211--252, 2015.

\bibitem{vaswani2017attention}
Ashish Vaswani, Noam Shazeer, Niki Parmar, Jakob Uszkoreit, Llion Jones,
  Aidan~N Gomez, {\L}ukasz Kaiser, and Illia Polosukhin,
\newblock ``Attention is all you need,''
\newblock {\em Advances in neural information processing systems}, vol. 30,
  2017.

\bibitem{mehta2021mobilevit}
Sachin Mehta and Mohammad Rastegari,
\newblock ``Mobilevit: light-weight, general-purpose, and mobile-friendly
  vision transformer,''
\newblock {\em arXiv preprint arXiv:2110.02178}, 2021.

\bibitem{yu2022metaformer}
Weihao Yu, Mi~Luo, Pan Zhou, Chenyang Si, Yichen Zhou, Xinchao Wang, Jiashi
  Feng, and Shuicheng Yan,
\newblock ``Metaformer is actually what you need for vision,''
\newblock in {\em Proceedings of the IEEE/CVF Conference on Computer Vision and
  Pattern Recognition}, 2022, pp. 10819--10829.

\bibitem{hu2018squeeze}
Jie Hu, Li~Shen, and Gang Sun,
\newblock ``Squeeze-and-excitation networks,''
\newblock in {\em Proceedings of the IEEE conference on computer vision and
  pattern recognition}, 2018, pp. 7132--7141.

\bibitem{tolstikhin2021mlp}
Ilya~O Tolstikhin, Neil Houlsby, Alexander Kolesnikov, Lucas Beyer, Xiaohua
  Zhai, Thomas Unterthiner, Jessica Yung, Andreas Steiner, Daniel Keysers,
  Jakob Uszkoreit, et~al.,
\newblock ``Mlp-mixer: An all-mlp architecture for vision,''
\newblock {\em Advances in Neural Information Processing Systems}, vol. 34, pp.
  24261--24272, 2021.

\bibitem{paszke2019pytorch}
Adam Paszke, Sam Gross, Francisco Massa, Adam Lerer, James Bradbury, Gregory
  Chanan, Trevor Killeen, Zeming Lin, Natalia Gimelshein, Luca Antiga, et~al.,
\newblock ``Pytorch: An imperative style, high-performance deep learning
  library,''
\newblock {\em Advances in neural information processing systems}, vol. 32,
  2019.

\bibitem{ramachandran2017searching}
Prajit Ramachandran, Barret Zoph, and Quoc~V Le,
\newblock ``Searching for activation functions,''
\newblock {\em arXiv preprint arXiv:1710.05941}, 2017.

\bibitem{51420}
Ilya Eckstein, James~Patrick Lee-Thorp, Joshua Ainslie, and Santiago Ontanon,
  Eds.,
\newblock {\em FNet: Mixing Tokens with Fourier Transforms}, 2022.

\bibitem{lin2014microsoft}
Tsung-Yi Lin, Michael Maire, Serge Belongie, James Hays, Pietro Perona, Deva
  Ramanan, Piotr Doll{\'a}r, and C~Lawrence Zitnick,
\newblock ``Microsoft coco: Common objects in context,''
\newblock in {\em European conference on computer vision}. Springer, 2014, pp.
  740--755.

\bibitem{carion2020end}
Nicolas Carion, Francisco Massa, Gabriel Synnaeve, Nicolas Usunier, Alexander
  Kirillov, and Sergey Zagoruyko,
\newblock ``End-to-end object detection with transformers,''
\newblock in {\em European conference on computer vision}. Springer, 2020, pp.
  213--229.

\end{thebibliography}

\end{document}